\documentclass{article}

\usepackage[scr=boondoxo,scrscaled=1.05]{mathalfa}
\usepackage[mathcal]{euscript}

\usepackage{mathtools}
\usepackage{lmodern}
\usepackage[short, nocomma]{optidef}
\usepackage{subcaption}
\usepackage{amsthm}
\usepackage{xcolor}

\usepackage{setspace}
\usepackage[margin=2cm]{geometry}
\usepackage{newtxtext}
\usepackage{newtxmath}
\usepackage[ruled, vlined,linesnumbered]{algorithm2e}
\usepackage[absolute]{textpos}

\usepackage{enumitem}
\usepackage{booktabs}
\usepackage{multirow}
\usepackage{array}
\usepackage[overload]{empheq}
\usepackage{bm}
\usepackage[noadjust]{cite}
\usepackage{makecell}

\DeclarePairedDelimiter{\norm}{\lVert}{\rVert}

\makeatletter
\newcommand*{\defeq}{\mathrel{\rlap{%
			\raisebox{0.3ex}{$\m@th\cdot$}}%
		\raisebox{-0.3ex}{$\m@th\cdot$}}%
	=}

\makeatletter
\def\thmheadbrackets#1#2#3{%
	\thmname{#1}\thmnumber{\@ifnotempty{#1}{ }\@upn{#2}}%
	\thmnote{ {\the\thm@notefont[#3]}}}
\makeatother

\newtheoremstyle{brakets}
{}
{}
{\normalfont}
{}
{\bfseries}
{.}
{ }
{\thmheadbrackets{#1}{#2}{#3}}

\newtheoremstyle{defbrakets}
{}
{}
{\normalfont}
{}
{\bfseries}
{.}
{ }
{\thmheadbrackets{#1}{#2}{#3}}

\newtheoremstyle{defproblem}
{}
{}
{\normalfont}
{}
{\bfseries}
{.}
{ }
{\thmheadbrackets{#1}{#2}{#3}}

\theoremstyle{brakets}
\newtheorem{thm}{Theorem}

\theoremstyle{defbrakets}

\newtheorem{defn}[thm]{Definition}

\theoremstyle{defproblem}
\newtheorem{plm}[thm]{Problem}
\newtheorem{rem}[thm]{Remark}

\makeatletter
\newcommand{\algrule}[1][.2pt]{\par\vskip.5\baselineskip\hrule height #1\par\vskip.5\baselineskip}
\makeatother

\newcommand{\z}[1]{\textcolor{black}{#1}}
\newcommand{\y}[1]{\textcolor{black}{#1}}

\let\oldnl\nl
\newcommand{\nonl}{\renewcommand{\nl}{\let\nl\oldnl}}

\begin{document}

	\title{\bfseries Learning Optimal Topology for Ad-hoc Robot Networks}
\author{Matin~Macktoobian\footnote{matin.macktoobian@mail.utoronto.ca}, Zhan Shu, and Qing Zhao}%
\date{Electrical and Computer Engineering Department\\ University of Alberta\\Edmonton, AB, Canada}
\maketitle

\begin{textblock}{14}(6,1)
	\noindent\textbf{\color{red}Published in 
		``IEEE Robotics and Automation Letter'' \\DOI: 10.1109/LRA.2023.3246845
		}
\end{textblock}

\begin{abstract}
In this paper, we synthesize a data-driven method to predict the optimal topology of an ad-hoc robot network. This problem is technically a multi-task classification problem. However, we divide it into a class of multi-class classification problems that can be more efficiently solved. For this purpose, we first compose an algorithm to create ground-truth optimal topologies associated with various configurations of a robot network. This algorithm incorporates a complex collection of optimality criteria that our learning model successfully manages to learn. This model is an stacked ensemble whose output is the topology prediction for a particular robot. Each stacked ensemble instance constitutes three low-level estimators whose outputs will be aggregated by a high-level boosting blender. Applying our model to a network of 10 robots displays over 80\% accuracy in the prediction of optimal topologies corresponding to various configurations of the cited network.
\end{abstract}

\textbf{keywords}: Networked robots, Model learning for control, Multi-robot systems, Machine learning for robot control

\maketitle
\doublespacing
\section{Introduction}
An ad-hoc robot network requires a setting of communicational routes between its robots using which the passage of information can be feasible. The overall characterization of robots of a network and their links involved in such routes is known as the topology of that network. A robot network shall constantly constitute a topology based on which its robots are, either directly or indirectly, connected to each other \cite{alsamhi2020blockchain,pennisi2015multi}. Topology dramatically impacts various aspects of communications in robot networks such as delay, fault tolerance, computation distribution, and connection reliability \cite{bertrand2013seeing}. Thus, finding the optimal topology, based on specific criteria, associated with a particular configuration of a robot network is often of utmost importance \z{in many applications such as rescue and search \cite{drew2021multi}, excavation \cite{caiazza2021application}, exploration \cite{macktoobian2013time}, etc.} 

Robot networks often enjoy high degrees of freedom associated with their robots which provide collective dexterity to perform their tasks. However, the cited rich dynamism must not jeopardize the connectivity of a their network topology \cite{batalin2004mobile}. Analytical syntheses of optimal topologies demand labor-intensive computations which are inefficient, should one take a massive network and/or complicated optimality criteria into account. In this regard, computational complexity of topology checking can exponentially grow in the case of particular topology formations \cite{xiangpeng2016unified,macktoobian2017optimal}, especially cycle topologies.

Chain topologies are a class of formations suitable for robot networks with a single actor robot while the remainder of its peers, except a base robot, solely contribute to the maintenance of the communication flow between the actor and the base robot \cite{li2014topology}. These topologies support limited types of missions. Instead, cycle topologies are legit options for various missions including surveillance, search and rescue, exploration/excavation, and so on. Nonetheless, \y{online computation of a cycle topology corresponding to a robot network essentially entails the computation of the largest cycle, known as backbone, available among the robots of that network. This problem is indeed equivalent to checking whether the graph of the network is Hamiltonian, which is an NP-hard problem \cite{mavrogiannis2021hamiltonian}. So, for a robot network with even a couple of robots, their optimal topology cannot be analytically achieved efficiently, assuming a set of optimality measures.}

\y{To relax the necessity of direct cycle topology computation, we alternatively seek the synthesis of a data-driven predictor trained by the data of various configurations of a robot network in addition to the optimal topology information associated with those configurations. When a motion step is planned for robots of a network, the result of those steps, as a new configuration, can be assessed by the topology predictor. If the predictor returns any (optimal) topology, then the maneuvers of the robots may be allowed. Otherwise, such a set of maneuvers shall be avoided as it would be highly probable that their execution leads to various detachments of the robots from their network.} To recapitulate, we seek the solution to the following problem.
\begin{plm}
	\label{plm}
	Suppose an algorithm, associated with a set of optimality criteria, using which one can yield the optimal topology corresponding to an ad-hoc robot network. Then, given a dataset comprising the ground-truth results of the presupposed algorithm, synthesize a data-driven model which can predict the optimal topology of an unknown configuration of that network.
\end{plm}
\subsection{Related Work}
\y{Various data-driven strategies have been employed to learn topological information of ad-hoc networks. As the first class of methods, convolutional graph networks exhibit remarkable potential to learn graph structures in fully-connected networks using convolutional graph layers \cite{chen2022learning,sarlin2020superglue}. However, these convolutional layers are not properly capable of learning topologies in incomplete graphs, which are common in general robot networks. These layers are also not ideal artifacts to systematically apply optimality constraints to learning models. Generative pre-training networks \cite{hu2020gpt} have been exceptionally successful in mimicking real topology data to construct similar fictitious ones. However, their advantages immediately vanish when one allows problem parameters to be continuous, similarly to the position signals in ad-hoc robot networks, instead of discrete.}

\y{Unsupervised learning strategies \cite{tscherepanow2010topoart} have also been used to perform learning processes without any pre-defined topological information. For example, in \cite{furao2006incremental}, an online incremental method was introduced in which agents are sequentially clustered, but the efficiency of this scheme for larger networks seems to be questionable. Moreover, due to the unsupervised nature of this method, it only minimizes a distance-based error term associated with network nodes.}

\y{Literature also includes topology learning strategies that are hybrid in terms of the usage of data structures. Namely, a sensor network topology learning approach \cite{marinakis2005learning} assumed the availability of a Markovian model describing the communicational interactions between its nodes. This model infers a topology based on the Monte Carlo maximization of the attachment of its nodes to each other. However, the strongly-restrictive assumption of this scheme, say, assuming agents' coordinates are all fixed, makes it practically inapplicable to robot networks. Learning network topologies based on computationally-affordable procedures have been studied, e.g., \cite{ghouti2013mobility}, at the cost of targeting sub-optimal topologies. Using reinforcement learning to converge to optimal topologies gave rise to slow and fluctuating convergence to other sub-optimal solutions \cite{testi2020blind}.}

\y{\subsection{Contributions}
	Given the review above, we elucidate our contributions to generally solving Problem \ref{plm}, as follows.
	\begin{enumerate}
		\item The general topology prediction for ad-hoc robot networks is a difficult multi-task classification problem in terms of achieving high prediction accuracies. So, we transform the problem to a set of simpler multi-class classification problems based on the divide-and-conquer paradigm. We embed the topological information of a robot network in a scalable manner such that the dimensions of the quoted multi-class classification problems are fully specified only based on the number of robots in a network. 
		\item Topology is generally a complex graph-based notion that cannot be easily expressed without graph data structures. Such graph data structures may not be processed in machine-learning pipelines as efficient as simple vectors of data. Thus, we partition a topology to a backbone cycle and a branch set, so that their information can be efficiently encoded to integer vectors. This approach preserves topological correlations of robots despite the transformation of the main multi-task classification problem to a set of multi-class ones associated with different robots. \z{The reason is that we apply no nonlinear transformation to robot data while partitioning the main multi-task classification problem to multiple multi-class classification problems. Thus, the resulting dataset represents unaltered characteristics of the intended robot network and correlations of robot location data.}
		\item The proposed learning model successfully predicts the results of the developed optimal topological label generator procedure which includes complex nonlinear structural optimality criteria. So, our learning model exhibits remarkable capacity to learn different optimality templates in deep representations of robot network data.
\end{enumerate}}
\y{To summarize, we collect spatial coordinates of the robots of a network and the data associated with the optimal topologies of some already-checked configurations of that network. Then, we generate local datasets each of which corresponds to one of the robots and its topological correlations with its peers. We synthesize a set of stacked ensembles of classifiers to solve multi-class classification tasks associated with all of the robots. This model is capable of predicting the optimal topology of a new unexplored configuration of that robot network given the coordinates of its robots in that configuration.}
\section{Methodology}
\label{sec:data}
\subsection{Ground-Truth Optimal Topology Establishment}
\label{subsec:ground}
The area around a typical robot may be radially partitioned into three different regions according to the quality of its communication with other peers which reside in those regions. In particular, communication quality of a robot is maximum with respect to its peers that are located at a distance not farther than a \textit{connectivity threshold} $\delta > 0$. The quality of communication then decreases when a robot communicates with a peer located outside of the aforesaid region but closer than the distance $\delta + \epsilon$, where $0<\epsilon<\delta$ is a \textit{tension bound factor}. Finally, the robot cannot communicate with its peers whose locations are beyond the boundary of the described region.       
\begin{defn}[Robot Network]
	Let $\delta$ and $\epsilon$ be connectivity threshold and tension bound factor, respectively, associated with a set of robots
	$
	\mathscr{R}\!\defeq\!\bigl\{\mathscr{R}_{i} \mathrel{\Big|} i\!\in\!\mathscr{I} \bigr\},
	$
	where $\mathscr{I}$ is an index set. Then, 3-tuple $\mathscr{N}\!\defeq\!(\mathscr{R}, \delta, \epsilon)$ represents a \textit{robot network}. We overload the notation $\mathscr{R}_{i}\!\defeq\!(x_{i}, y_{i})$ to also denote the planar coordinate of robot $\mathscr{R}_{i}$.
\end{defn}
The two definitions below specify the robots that are candidates for reliable connections or critical ones to a peer. Critical connections are those whose quality is not ideal.
\begin{defn}[Reliable Set]
	Let $\mathscr{N}\!=\!(\mathscr{R}, \delta, \epsilon)$ be a robot network. Then, given a robot $\mathscr{R}_{i} \in \mathscr{R}$, the \textit{reliable set} $\mathscr{R}_{i}^{+}$ is defined as follows.
	\begin{equation}
		\mathscr{R}_{i}^{+} \defeq \bigl\{r \in \mathscr{R}\!\setminus\!\{\mathscr{R}_{i}\} \mathrel{\big|} \norm{\mathscr{R}_{i} - r} \le \delta \bigr\}
	\end{equation}
\end{defn}
\begin{defn}[Critical Set]
	Let $\mathscr{N} = (\mathscr{R}, \delta, \epsilon)$ be a robot network. Then, given a robot $\mathscr{R}_{i} \in \mathscr{R}$, the \textit{critical set} $\mathscr{R}_{i}^{-}$ is defined as follows.
	\begin{equation}
		\mathscr{R}_{i}^{-} \defeq \bigl\{r \in \mathscr{R}\!\setminus\!\{\mathscr{R}_{i}\} \mathrel{\big|} \delta < \norm{\mathscr{R}_{i} - r} \le \delta+\epsilon \bigr\}
	\end{equation}
\end{defn}
We define the auxiliary notions of link set, associated with a robot network, and connection relation between two robots, as below.
\begin{defn}[Link Set]
	Given a robot network $\mathscr{N}$, its \textit{link set} reads as the following\footnote{\z{Symbols $\wedge$ and $\dot{\cup}$ represent logical conjunction and disjoint union, respectively.}}:
	\begin{equation}
		\begin{split}
			\mathscr{L}_{\mathscr{N}} \defeq \bigl\{ (x,y) \mathrel{\big|} \bigl(\forall i,j \in \mathscr{I}\bigr)  \bigl[& x \in \bigl(\mathscr{R}_{i}^{+} \dot{\cup}\mathscr{R}_{i}^{-}\bigr) \wedge \\& y \in \bigl(\mathscr{R}_{j}^{+} \dot{\cup}\mathscr{R}_{j}^{-}\bigr)\bigr] \bigr\}.
		\end{split}
	\end{equation}
\end{defn}
Based on these definitions, the graph of a robot network is formalized as follows. 
\begin{defn}[Robot Network Graph]
	Let $\mathscr{N} = (\mathscr{R}, \delta, \epsilon)$ be a robot network. Then,  $\mathscr{G}_{\mathscr{N}} \defeq (V_{\mathscr{N}}, E_{\mathscr{N}})$ is the \textit{robot network graph} associated with $\mathscr{N}$ such that $
	V_{\mathscr{N}} \defeq \mathscr{R}$ and $ E_{\mathscr{N}} \defeq \mathscr{L}_{\mathscr{N}}$ are the set of vertices and edges of the graph, respectively.
\end{defn}
Now, we establish the definition of a backbone cycle of a robot network which is one of the two elements shaping the definition of a robot network topology, as we will later see.
\begin{defn}[Backbone Cycle\protect\footnote{In a graph, a cycle is a closed path of successive vertices and links in which neither vertices (except the start/end one) nor links may be repeated.}]
	Let $\mathscr{G}_{\mathscr{N}}$ be the graph of a robot network $\mathscr{N}$. Then, $\psi_{\mathscr{N}} \defeq (V_{\psi_{\mathscr{N}}}, E_{\psi_{\mathscr{N}}})$ is a \textit{backbone cycle} of $\mathscr{G}_{\mathscr{N}}$ such that
	$V_{\psi_{\mathscr{N}}} \subseteq \bigl\{i \in \mathscr{I} \mathrel{\big|} \mathscr{R}_{i}^{+} \bigr\}.$	
\end{defn}
The relation below paves the way for scoping the communicational reachability of a robot in a backbone cycle by another peer that is not an immediate neighbor of it outside of the backbone cycle.
\begin{defn}[Indirect Reachability]
	If robot $\mathscr{R}_{i}\!\in\!\mathscr{R}$ can communicationally reach a backbone cycle node $v$ via a succession of its neighboring vertices excluding node set $W\!\subset\!\mathscr{R}\!\setminus\!\{\mathscr{R}_{i}, v\}$, then we say $\mathscr{R}_{i}$ \textit{indirectly reaches} $v$, i.e., $\mathscr{R}_{i} \stackrel{W}{\mathrel{\leadsto}}  v$. 
\end{defn}
The second element of a robot network topology is a branch set defined as follows.
\begin{defn}[Branch]
	Given a robot network $\mathscr{N}$ and a backbone cycle node $v\in V_{\psi_{\mathscr{N}}}$, $b_{v} \subseteq \{ \mathscr{R}_{i}\mathrel{\big|}\mathscr{R}_{i}\!\in \mathscr{R}\!\setminus\!V_{\psi_{\mathscr{N}}}\}$ is a branch, with respect to $v$, if $b_v$ includes no cycles and $\bigl(\forall \mathscr{R}_i\in b_{v}\bigr)\mathscr{R}_{i} \stackrel{V_{\psi_{\mathscr{N}}}}{\mathrel{\leadsto}} v$.
\end{defn}
\begin{defn}[Robot Network Topology]
	Let $\mathscr{I}$ be an index set whose cardinality determines the number of the robots in its associated robot network $\mathscr{N}$. Given, the set of all branches $\mathscr{B}_{\mathscr{N}}\!\defeq\!\bigl\{b_{v}\bigr\}_{v \in V_{\psi_{\mathscr{N}}}}$, the \textit{robot network topology} $\mathscr{T}_{\mathscr{N}} \defeq (\psi_{\mathscr{N}}, \mathscr{B}_{\mathscr{N}})$ corresponds to $\mathscr{N}$ if both of the following conditions are simultaneously fulfilled.
	\begin{subequations}
		\begin{align}[left ={ \empheqlbrace}]
			& V_{\mathscr{B}_{\mathscr{N}}} \cup V_{\psi_{\mathscr{N}}} = V_{\mathscr{N}} \label{eq:topo_1}\\
			&\bigl(\forall v \not\in V_{\psi_{\mathscr{N}}}\bigr) \bigl[\bigl(\exists b \in \mathscr{B}_{\mathscr{N}}\bigr) v \in b\bigr] \label{eq:topo_2}
		\end{align}
	\end{subequations}
\end{defn}
\begin{rem}
	Equation (\ref{eq:topo_1}) mandates that the unification of the robots belonging to the backbone cycle and all of the branches must cover all of the robots of the network. Equation (\ref{eq:topo_2}) requires that any robot that does not belong to the backbone cycle must belong to one of the branches of the topology.
\end{rem}
The auxiliary notions of connection relation and robot degree, defined below, are the last required pieces to formulate optimality for a robot network topology.
\begin{defn}[Connection Relation]
	\textit{Connection relation} $\mathscr{C}(\mathscr{R}_i,\mathscr{R}_j)$ holds if robots $\mathscr{R}_{i}$ and $\mathscr{R}_{j}$ are connected via a link.
\end{defn}
\begin{defn}[Robot Degree]
	Given a robot $\mathscr{R}_{i}$, its \textit{degree}, denoted by $\mathscr{D}_{i}$, is defined as the cardinality of the set
	$
	\bigl\{\forall \mathscr{R}_j \in \mathscr{R}\!\setminus\!\mathscr{R}_{i} \mathrel{\big|}\mathscr{C}(\mathscr{R}_i,\mathscr{R}_j)\bigr\},
	$
	which refers to the number of the connections of $\mathscr{R}_i$ to other robots.
\end{defn}
\begin{defn}[Optimal Robot Network Topology]
	\label{defn:ORNT}
	A robot network topology is \textit{optimal}, represented by $\mathscr{T}_{\mathscr{N}}^{\star} \defeq (\psi_{\mathscr{N}}^{\star}, \mathscr{B}_{\mathscr{N}}^{\star})$, if the following conditions are met: (i) $\psi_{\mathscr{N}}^{\star}$ is the largest cycle of $\mathscr{N}$; (ii) All branches of $\mathscr{B}_{\mathscr{N}}^{\star}$ are minimum spanning trees, with respect to the distance between robots, over their robot sets;
	(iii) Links associated with every branch in $\mathscr{B}_{\mathscr{N}}^{\star}$ are as reliable as possible\footnote{The more reliable connections are embedded into a topology, the more reliable it is.};
	(iv) Links corresponding to every branch in $\mathscr{B}_{\mathscr{N}}^{\star}$ are as structurally-distributed as possible, i.e., given an arbitrary $b \in \mathscr{B}_{\mathscr{N}}^{\star}$, for all $\mathscr{R}_{i}, \mathscr{R}_{j} \in b$ such that\footnote{Unary operator $|\cdot|$ denotes the absolute value of its argument.} $\mathscr{C}(\mathscr{R}_i,\mathscr{R}_j)$, $|\mathscr{D}_{i} - \mathscr{D}_{j}|$ has to be minimum.
\end{defn}
\begin{rem}
	\z{The topology defined above is optimal in the sense that it minimizes congestion risks and maximizes link reliability against communication faults and noise. In particular, the requirements 1, 2, and 4 intend to realize the maximum connectivity in a network with the minimum number of branches and connections. So, the message passing between robots shall be minimized that contributes to the minimization of congestion risks in such networks. The requirement 4 particularly spreads the link distribution of the optimal topology as much as possible so that no robot acts like a hub node. A hub node, to which many robots are connected, makes the topology more vulnerable to faults because if it becomes non-functional, all of the nodes connected to it may potentially lose their connectedness to the network, as well. The requirement 3, on the other hand, maximizes the number of reliable links, instead of critical ones, to minimize the vulnerability of links against noise and faults.}
\end{rem}
\subsection{Topology Computation Algorithm}
\label{subsec:alg}
In this section, we develop an algorithm, i.e., Algorithm \ref{alg:OTC}, to compute the optimal topology of a typical robot network. This algorithm particularly provides the ground-truth knowledge to be later used for dataset preparation and predictor synthesis to solve Problem \ref{plm}.

One needs to feed a set of inputs to the algorithm: first, an index set which distinguishes robots of a robot network; second, the configuration parameters of a network which are its zone range, connectivity threshold, and tension bound factor; and finally a random generator which uniformly generates the locations of the network's robots. In line 1, a variable is created to contain the robots of the network. Namely, we assign two random numbers, as a planar coordinate, to each robot in line 2, followed by a location validation in line 3. Here, if the generated coordinate provides the robot (at least critical) connectedness with respect to at least one of its neighboring peers, then we add the robot to the robot network in line 4. Otherwise, the coordinate is discarded, in line 5, and a new one has to be generated for the robot. Once all robots correspond to communicationally valid coordinates, reliable and critical sets associated with them are computed, in line 6. In line 7, the robots are sorted based on the cardinalities of their reliable sets. Any equality of those measures between any pairs of robots makes the cardinalities of their critical sets as a tie-breaker in the course of the quoted sorting process. In the case of the equality of these factors, robots are randomly placed in the current index of the sorted set. Then, the optimal robot network topology may be achieved by computing its constituents, i.e., its backbone cycle (ine 8) and its branch set (line 9). In the end, we yield the desired optimal network topology, in line 10.  
\begin{algorithm}[t]
	\caption{Optimal Topology Computer}
	\label{alg:OTC} 
	\SetKwInput{KwData}{Inputs}
	\SetKwInput{KwResult}{Output}
	\KwData{Robot network index set $\mathscr{I}$\\  
		\hspace*{13.4mm}Zone range $\mathscr{Z}$\\
		\hspace*{13.4mm}Connectivity threshold $\delta$\\
		\hspace*{13.4mm}Tension bound factor $\epsilon$\\
		\hspace*{13.4mm}Uniform random generator $U(0,\mathscr{Z})$} 
	\KwResult{The optimal topology $\mathscr{T}_{\mathscr{N}}^{\star}$ of a robot network}
	\algrule[1pt]
	$\mathscr{R} \leftarrow \varnothing$\\
	\nonl\ForEach{$i \in \mathscr{I}$}{
		$\mathscr{R}_{i} \leftarrow (U(0,\mathscr{Z}), U(0,\mathscr{Z}))$\\
		\If{$\mathscr{R}_{i}$ is communicationally connected to at least one of the robots already generated}{$\mathscr{R} \leftarrow \mathscr{R} \dot{\cup}\mathscr{R}_{i}$}
		\nonl\Else{Regenerate $\mathscr{R}_{i}$}
		Compute $\mathscr{R}_{i}^{+}$ and $\mathscr{R}_{i}^{-}$
	}
	$\mathscr{R} \leftarrow $ Sort $\mathscr{R}$ based on the cardinalities of $\mathscr{R}_{i}^{+}$ and \hspace*{7mm}$\mathscr{R}_{i}^{-}$ according to their coordination priorities \hspace*{7mm}$(\forall i \in \mathscr{I})$\\
	\nonl Given $\mathscr{N} = (\mathscr{R}, \delta, \epsilon)$ and considering the conditions of Definition \ref{defn:ORNT}:\\	
	~~~~~~~ Compute $\psi^{\star}_{\mathscr{N}}$ based on the condition (i)\\
	~~~~~~~ Compute $\mathscr{B}^{\star}_{\mathscr{N}}$ fulfilling the conditions (ii), (iii) \hspace*{8mm}(if not violating (ii)), and (iv) (if not violating \hspace*{8mm}(ii) and (iii))\\
	\Return $\mathscr{T}_{\mathscr{N}}^{\star} \leftarrow (\psi^{\star}_{\mathscr{N}},\mathscr{B}^{\star}_{\mathscr{N}})$
\end{algorithm}
\begin{figure}
	\centering\includegraphics[scale=0.47]{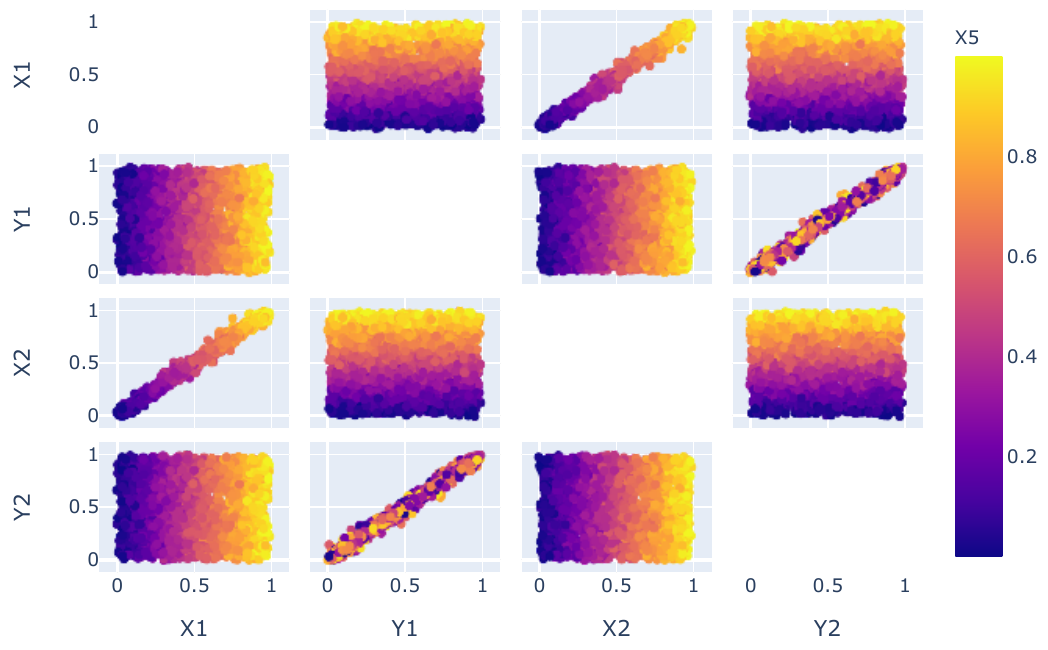}
	\caption{Interrelated correlations among robot coordinates in view of each other's classifiability. (\z{The axes of each subfigure are absolute correlation factors.} For example, the scattering dependency of the classifiability of the horizontal coordinate component of robot 5, i.e., X5, is displayed with respect to the coordinate components of robots 1 and 2. One can observe that if the coordinates of a particular robot, say, X1-Y1, are taken into account, the separation boundary to classify X5 is different than the correlated coordinate cases, such as X1-Y2 or X2-Y1. In these cases, the classifiability of X5 exhibits more convoluted trends.)}
	\label{fig:corr}
\end{figure}

\z{The computational bottleneck in Algorithm 1 corresponds to line 8 in which the backbone cycle has to be computed. The computational complexity of this NP-hard problem, as stated in the introduction, is of factorial order. This complexity is related to the dataset generation phase. So, one may argue that for large-scale networks, one has to train the model in an offline manner. In smaller networks, online training may be possible depending on available computational resources. Once a dataset, thereby its corresponding trained model, is available, one can predict the optimal topology in real-time.}
\subsection{Topology Embedding and Dataset Formation}
\label{subsec:embed}
We embed the minimum data corresponding to various optimal topology scenarios, computed by Algorithm \ref{alg:OTC}, in a dataset for the purpose of topology predictions. In this regard, robot locations of an intended network are key enablers to track topology variations from one configuration of the robot network to another. So, given an index set $\mathscr{I}$ associated with a robot set $\mathscr{R}$, the location information of $\mathscr{R}$ is denoted by row-vector 
$
\Xi^{d} \defeq 
\begingroup
\setlength\arraycolsep{3pt}
\begin{bmatrix}
	\mathscr{R}^{x}_{1} & \mathscr{R}^{y}_{1} & \cdots & \mathscr{R}^{x}_{|\mathscr{I}|} & \mathscr{R}^{y}_{|\mathscr{I}|}
\end{bmatrix}_{1 \times 2|\mathscr{I}|},
\endgroup
$ where $\mathscr{R}^{x}_{(\cdot)}$ and $\mathscr{R}^{y}_{(\cdot)}$ are the $x$- and $y$-component of the $(\cdot)$th robot. We also have to embed the topological information of the algorithm's output as the labels associated with a robot network configuration in $\Xi^{t}$. Namely, we assign each robot $\mathscr{R}_{i}$ to a cluster $\Xi^{t}_{i}$ such that the following conditions hold. 
\begin{subequations}
	\label{Phi:def}
	\begin{align}[left ={ \empheqlbrace}]
		& \bigl(\forall \mathscr{R}_{i} \in \mathscr{R}\bigr) \bigl[\bigl(\mathscr{R}_{i} \in V_{\psi_{\mathscr{N}}}\bigr) \Rightarrow \bigl(\Xi_{i}^{t} \defeq i \bigr)\bigr]\\
		& \bigl(\forall \mathscr{R}_{i} \in \mathscr{R}\bigr)\bigl[ \bigl\{\bigl(\mathscr{R}_{i} \not\in V_{\psi_{\mathscr{N}}}\bigr) \wedge\bigl(\exists \mathscr{R}_{j} \in V_{\psi_{\mathscr{N}}} \bigr)\nonumber\\&\hspace*{1.8cm}\bigl(\mathscr{R}_{i} \stackrel{V_{\psi_{\mathscr{N}}}}{\mathrel{\leadsto}}\mathscr{R}_{j}\bigr)\bigr\}\Rightarrow \bigl(\Xi_{i}^{t} \defeq j \bigr)\bigr] 
	\end{align}
\end{subequations}
In other words, if a robot belongs to the backbone cycle of its topology, its cluster entry equals its index. Otherwise, its index is that of the backbone-cycle robot that is indirectly reachable by that robot. Thus, an augmented data record $\Xi\defeq 
\begin{bmatrix}
	\Xi^{d} \mid \Xi^{t}
\end{bmatrix}_{1 \times 3|\mathscr{I}|}
$ may be obtained. Finally, the overall dataset is a $(n\!\times\!3|\mathscr{I}|)$ matrix including $n$ records of $\Xi$ associated with different configurations of its corresponding robot network.
\subsection{Optimal Topology Predictor Synthesis}
\label{subsec:synthesis}
\subsubsection{Approach}
\label{subsubsec:method}
The problem of optimal topology prediction for a robot network may be transformed to a classification problem, in which each robot shall be assigned to a particular cluster associated with the overall clusters of its network topology. The dataset structure, introduced in Section \ref{subsec:alg}, implies that cluster values are non-binary and their majority equals the number of robots in a desired robot network. Thus, this problem is inherently a  multi-task classification problem which is notoriously difficult when the number of labels and their possible values are both relatively large. A multi-task problem based on such a noticeable number of features may require a massive number of records to cover various permutations of all those integer entries so that under-fitting would be less likely to happen. Alternatively, we take a divide-and-conquer strategy into account to transform a multi-task topology prediction of a robot network to a set of multi-class classification problems each of which corresponds to the cluster prediction for only one of the robots of that network.

Based on this divide-and-conquer strategy, we replicate the main dataset of the multi-task problem, as many as the number of the robots in a network, such that each replicated dataset only includes the target label columns of the main dataset, but the input data of the main dataset is totally copied to all replicated datasets. We preserve the total input data of the main dataset in all replica datasets because a topology synthesis process has to consider the correlations between the robots of the network. In particular, Fig. \ref{fig:corr} shows a sample set of correlations between coordinate components of two typical robots. Predictions of all those distributed classifiers collectively represent the solution to the problem.

Ensemble learning is a technique to exploit the prediction power of various types of estimators to predict a particular feature. The more complex a machine learning problem is, the more significant the impact of ensemble learning will be. Thus, we propose OpTopNET, a particular network of stacked ensembles of multi-class classifiers each of which performs the optimal topology prediction corresponding to one of the multi-class classification problems described above. 
\subsubsection{Architecture}
\label{subsubsec:arch}
\begin{figure*}
	\centering
	\begin{subfigure}{0.73\textwidth}
		\centering
		\includegraphics[width=1.0\linewidth]{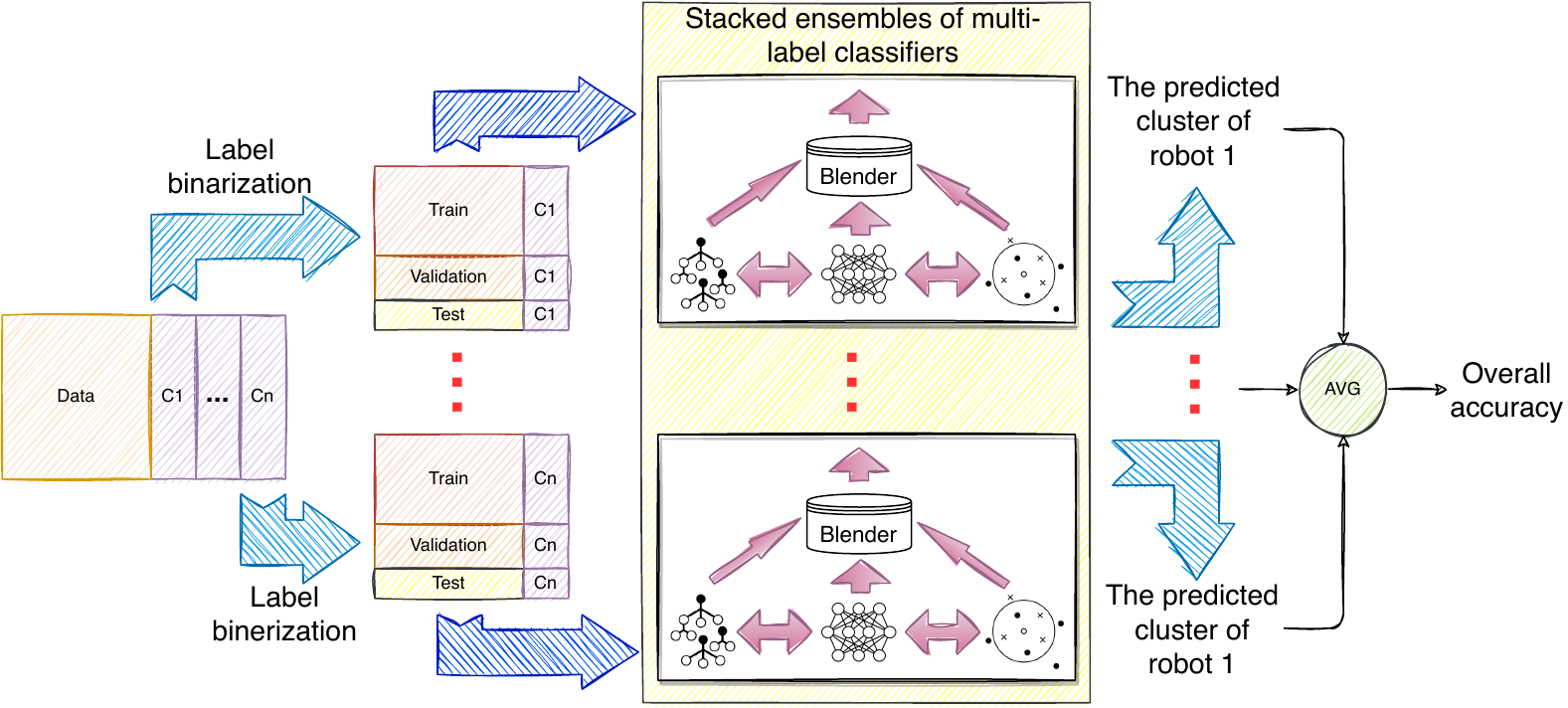}
		\caption{The architecture of OpTopNET}
		\label{fig:arch}
	\end{subfigure}%
	\hspace*{0cm}\begin{subfigure}{0.25\textwidth}
		\centering
		\includegraphics[width=1.\linewidth]{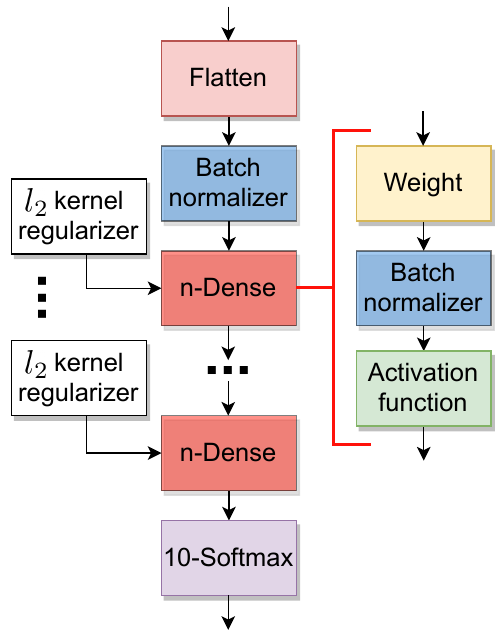}
		\caption{The structure of the deep neural network in OpTopNET.}
		\label{fig:deform}
	\end{subfigure}
	\caption{Learning model specification}
	\label{fig:layer}
\end{figure*}

The architecture of OpTopNET is depicted in Fig. \ref{fig:arch}. This computational pipeline first replicates the main dataset to $n$ local instances each of which includes the full input data associated with the locations of robots. However, they differ in that each one comprises the cluster labels of only one robot of that network. The label IDs are inherently integers, nonetheless there is no semantic relation between them. To address this issue, binarized versions of label IDs of the main dataset are taken into account in the local replicas. Then, each local dataset is used to train a particular stacking ensemble of classifiers. In each instance of these aggregated classifiers, the local dataset is, in parallel, fed into three diverse types of classifiers such as a random forest classifier, a $k$-neighbors classifier, and a deep neural network classifier. This type selection seeks a sufficiently-diverse set of classifiers whose error types are potentially very different. Thus, this diversity of error types may increase the overall stacked ensemble's accuracy. Because of the geometrical nature of the topology prediction, the $k$-neighbor classifier specifically contributes to the consideration of the metric distances between various topologies in the course of predictions \cite{macktoobian2021data}. The planned random forest classifier is used for better stable decision-making in complex scenarios such as those including large backbone cycles. Finally, the embedded deep network classifier seeks effective syntheses of deep representations for large ad-hoc robot networks whose dynamics may not be fully captured by the former classifier types. One may increase the number of the participating classifiers, but in practice, it not only makes training phase longer to be completed but also may not noticeably improve the accuracy of predictions.

Given a classification step, predictions of the cited low-level classifiers are employed to train a XGBoost-based blender deep network at the top of each stacked ensemble. Since each high-level blender collects and aggregates the predictions of its low-level classifying peers in the course of a classification task, one expects that the blender's prediction outperforms those of the low-level classifiers. The output of each stacked ensemble is a predicted cluster associated with the robot which is represented by that ensemble. So, the set including all those local cluster predictions constitutes the overall predicted optimal topology of the network of those robots. Moreover, the overall accuracy of OpTopNET is the average accuracy of its underlying stacked ensembles. The rationale to consider XGBoost blenders, instead of other options such as CatBoost, in the OpTopNET architecture is twofold. First, regularization and cross-validation are directly applicable to its training phases. So, one enjoys more degrees of freedom to further tune the ensemble beyond tuning hyperparameters of its constituent classifiers. Second, its effective tree pruning provides variable feature importance that is a highly-desirable characterization for a blender. 

The planned $k$-neighbors classifier, random forest classifier, and XGBoost blender are standard artifacts of Scikit-Learn library whose optimally-tuned hyperparameters are specified in Section \ref{subsec:res}. The deep neural network models of OpTopNET can be efficiently formulated by the sequential models of Keras library. The internal structure of the employed deep neural network is rendered in Fig. \ref{fig:deform}. Namely, a series of dense layers are sequentially connected to each other each of which is batch normalized. We also further regularize the model by adding $l_2$ kernels to all of its dense layers. A softmax layer, including 10 neurons, provides the output for each multi-class classification subproblem. 

\z{The applicability of an OpTopNET model associated with a network to another network depends on the identicality two factors in both networks: the number of robots in those networks, and their definitions of optimal topology. If one varies the number of robots from one network to another, then the number of the dataset columns of the first network is not compatible with the number of the robots of the second network. So, if one removes or add another robot from/to a network, its corresponding dataset has to be regenerated, and a new OpTopNET model has to be synthesized. Furthermore, if one assumes different optimality criteria for two networks, then their datasets have to be generated separately based on their specific optimal topology definitions. But, if these two factors are the same for two networks, one can train the model according to the data of one network and use it to predict optimal topologies for another one.}

\z{One may note that the robotic context to our approach is particularly general. We only consider a set of general robotic assumptions such as mobility and dynamism of robots and their telecommunication capabilities (via bluetooth or any other protocol). This generality expands the scope of the usage of our approach in various applications of ad-hoc robot networks}
\section{Simulations}
\label{sec:sim}
\subsection{Dataset Characterization and Setup}
\label{subsec:char}
Our dataset includes 2000 records of various configurations of a 10-robot\footnote{\y{In many real-world applications \cite{huang2022construction,hu2021decentralized}, the number of robots is around 10 to simultaneously exploit their cooperations and minimize complexities such as safety and collision avoidance concerns.}} network as well as their optimal topologies, computed by Algorithm \ref{alg:OTC}, in the format presented in Section \ref{subsec:embed}. We assign a value of 1 to the zone range, while values corresponding to connectivity threshold and tension bound factor are 0.5 and 0.1, respectively. We first shuffle the dataset before forking the local datasets associated with each of the multi-class classification problems. As already explained in Section \ref{subsubsec:arch}, each local dataset is associated with the target labels of a particular robot of the network. The label are indeed integer cluster IDs which are encoded to binary vectors. We also decrease the index of each robot by one unit so that the ID range [1,10] is substituted with [0,9] to be compatible with the numbering convention of output layers of deep neural networks under Keras framework.

\y{We take separate disjoint splits of local datasets into account for training, validation, and testing purposes of OpTopNET estimators, say, 72\%, 18\%, and 10\% of the overall number of the available records, respectively.}
\y{We take snapshots of network states, known as configuration, as the inputs of Algorithm \ref{alg:OTC}. The collision-free navigation of robots from one configuration to another is governed by a meta navigation function \cite{macktoobian2022meta}.}
\subsection{Results\protect\footnote{The simulations are performed on a Windows 10 x64 machine supported by a Core i7 1.80 GHz processor, 8GB RAM, and an Intel UHD Graphics 620. The following specific libraries are employed in the course of all performed simulations on Python 3.7.3: Tensorflow and Keras 2.5.0, Scikit-Learn 0.24.2.}\protect\footnote{The implementation of the dataset generator routine, i.e., Algorithm \ref{alg:OTC}, by which the dataset of this study is generated, may be found in https://git.io/JXKSb.}}
\label{subsec:res}
\begin{table}[b]
	\centering\caption{Hyperparameter settings and performance report associated with the synthesized OpTopNET.}
	\label{tbl:res}
	\hspace*{-16mm}
	\begin{tabular}{clccccccccccc}
		\toprule  
		\multirow{2}{*}[0.0em]{\bfseries\makecell{Classifier \\ type}} &\multirow{2}{*}[0.0em]{\bfseries\makecell{optimal parameters \\and measures}}&\multicolumn{10}{c}{\bfseries Cluster classification accuracy per robot}&\multirow{2}{*}[0.0em]{\bfseries\makecell{Average\\accuracy (\%)}}\\\cmidrule{3-12}
		&&1&2&3&4&5&6&7&8&9&10&\\
		\cmidrule{1-13}
		\multirow{3}{*}[-0.7em]{\makecell{Random Forest}} & Number of estimators &\multicolumn{10}{c}{100}&\multirow{3}{*}[-0.5em]{76.2} \\\cmidrule{3-12}& Depth & \multicolumn{10}{c}{5}\\\cmidrule{3-12}&Accuracy (\%)&89.0&90.0&88.5&85.0&75.0&74.5&72.5&61.5&64.0&62.0\\\cmidrule{1-13}
		\multirow{5}{*}[-1.4em]{$k$-Neighbors} & Number of neighbors & 6 & 5 & 9&9&9&9&9&9&9&9 & \multirow{5}{*}[-1em]{76.4}\\\cmidrule{3-12} & Weights & \multicolumn{10}{c}{Uniform} & \\\cmidrule{3-12} & Algorithm & Ball tree & Brute & Auto&Auto&Auto&Auto&Auto&Auto&Auto&Auto &\\\cmidrule{3-12} & p& 3 & 1 & 2&2&2&2&2&2&2&2 &\\\cmidrule{3-12}& Accuracy (\%) &88.5&93.5&85.0&81.5&76.0&76.5&72.5&64.5&66.5&59.0&\\\cmidrule{1-13}
		\multirow{4}{*}[-1.1em]{\makecell{Deep Neural \\ Network}} & Number of hidden layers & 3&3&3 & 1 & 3& 4& 3&3&3 & 4& \multirow{4}{*}[-0.8em]{76.7}\\\cmidrule{3-12}& Number of neurons &149&255&219&145&236&284&296&136&190&239\\\cmidrule{3-12} & Learning rate & \multicolumn{10}{c}{0.0001}\\\cmidrule{3-12}& Accuracy (\%) &89.7&91.1&89.5&80.2&78.2&76.2&70.0&65.9&64.1&62.0\\\cmidrule{1-13}
		\multirow{5}{*}[-1.5em]{\makecell{Stacked XGBoost\\Blender}} & Number of estimators & 100 & 100 & 130 & 130 & 130& 180& 180&180&180&180& \multirow{5}{*}[-1em]{81.3}\\\cmidrule{3-12}& Max depth & 4&3&4&5&3&4&4&3&4&5&\\\cmidrule{3-12}& Booster & \multicolumn{10}{c}{gbtree}& \\\cmidrule{3-12}&Learning rate & \multicolumn{10}{c}{0.0002}\\\cmidrule{3-12}&Accuracy (\%) &92.4 & 94.0 & 91.3 & 86.1 & 82.1& 80.2& 75.2& 74.1& 68.1 & 69.0&\\
		\bottomrule		
	\end{tabular}
\end{table}
In the course of the optimization of the hyperparameters of OpTopNET, selected batch size is 32. In the case of random forest classifiers and $k$-neighbors classifiers, we use 10 splits of input data to randomly spot the hyperparameters that maximize the accuracy metrics of those estimators. In particular, we sweep the number of estimators and the depth corresponding to random forest classifiers in integer ranges [100, 500] and [3, 10], respectively. For $k$-neighbors classifiers, number of neighbors and the power parameter P are swept within integer ranges [1, 10] and [1, 3], respectively. Algorithm hyperparameter options are ball tree, kd tree, brute, and auto. Moreover, we assess uniform and distance options for the weights hyperparameter. After performing randomized cross-validations, as Table \ref{tbl:res} expresses, the average accuracy performances of the optimally-tuned random forest classifiers and $k$-neighbors classifiers are 76.2\% and 76.4\%, respectively.

Deep neural network elements of OpTopNET are initialized based on the following specification. Batch normalization momentum associated with the batch normalization layers is 0.999. We take two strategies into account to properly regularize our model. First, for each deep neural network, we apply an $l_2$ regularization factor of 0.01 to every dense layer of that model. Second, we schedule the learning rate dynamics of that model based on the exponential decay profile
$
\eta(t) \defeq \eta_{0}\cdot 10^{-t/s},
$ where $s$ is the number of steps per epoch.

We employ scaled exponential linear unit as the activation function of each deep neural network model, whose kernel is initialized by lecun\_normal method to maximize the compatibility between each pair of activation function and kernel initializer. Our Adam optimizer is supported by momentum decay parameter $\beta_{1}\!=\!0.9$ and scaling decay parameter $\beta_{2} = 0.999$. A sparse categorical cross entropy is considered as the loss function of each deep neural network model. We perform cross-validation over 5 folds of data within 10 iterations for 90 epochs within which hyperparameter optimization is based on accuracy metric. In the course of the hyperparameter tuning, the number of hidden layers may very between 1 to 4. Number of the neurons in each dense layer can be any integer between 1 and 300. The lower-bound and the upper bound of the initial learning rate are 0.0001 and 0.03, respectively. Table \ref{tbl:res} illustrates the found optimal hyperparameters as well as the overall average accuracy of the optimally-tuned deep neural network which is 76.7\%.

\begin{figure}
	\centering\includegraphics[scale=0.35]{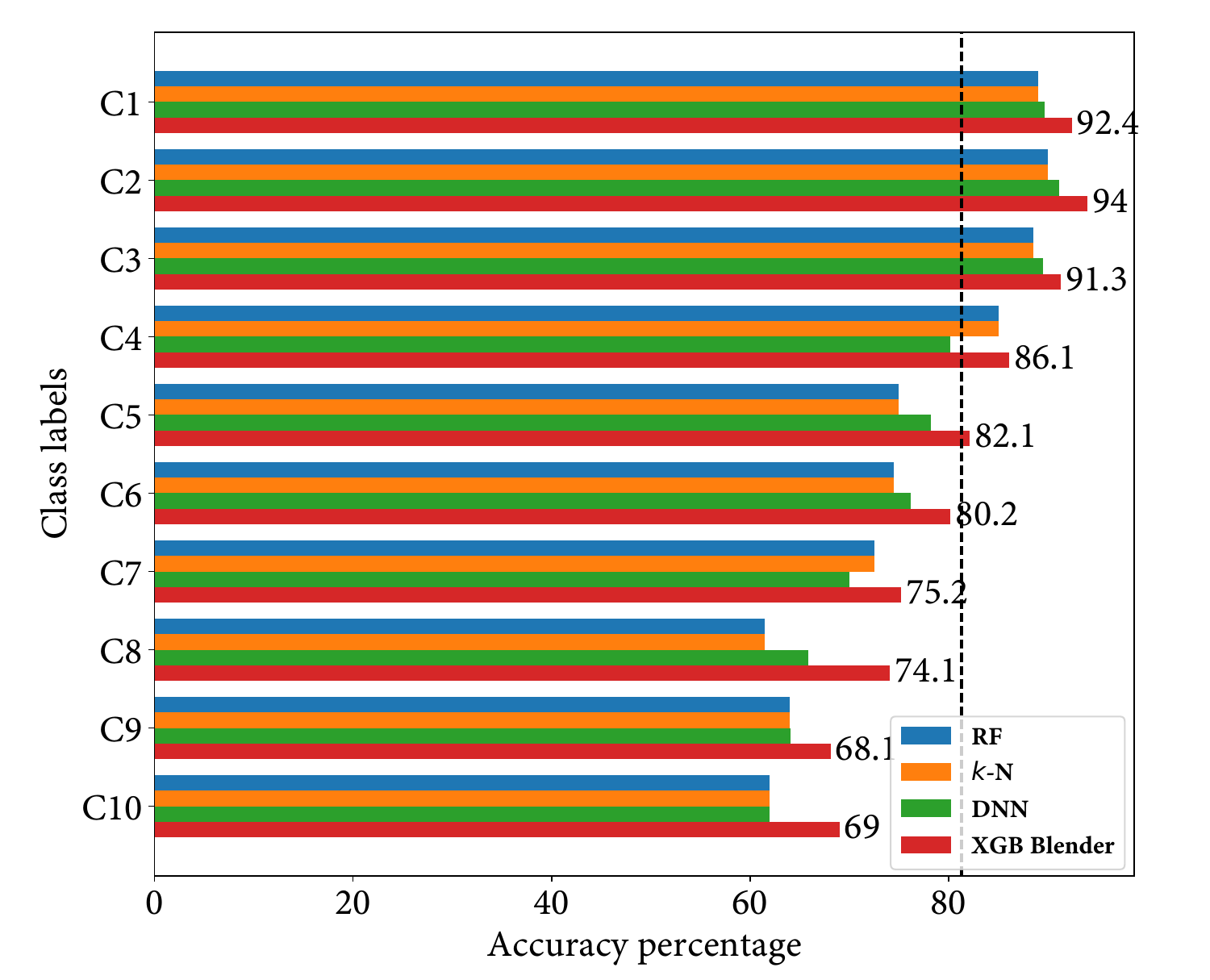}
	\caption{The average classification accuracy of ensemble classifiers and those of their constituent classifiers. (The vertical dashed line represents the overall average accuracy of the ensembles.)}
	\label{fig:acc}
\end{figure}

Finally, we construct stacked ensembles of each multi-class classifier to boost the performance of the single estimators investigated above. For this purpose, we use the XGBoost library, each instance of which is fed by the optimally-tuned estimators of a specific multi-class classification sub-problem. Max depth hyperparameter may vary between 3 and 10. Learning rate and gbtree instance ranges are from 0.0001 to 0.5. We then cross-validate the ensemble classifiers over 5 folds of data in the course of 10 iterations. The results\z{, which solely correspond to the test data,} are reflected in Table \ref{tbl:res} indicating that the overall average accuracy of all stacked ensembles are boosted to 81.3\%. Accordingly, Fig. \ref{fig:acc} specifically demonstrates how the XGBoost-based blenders outperform their low-level estimators in terms of accuracy. 
\begin{figure}
	\centering\includegraphics[scale=0.38]{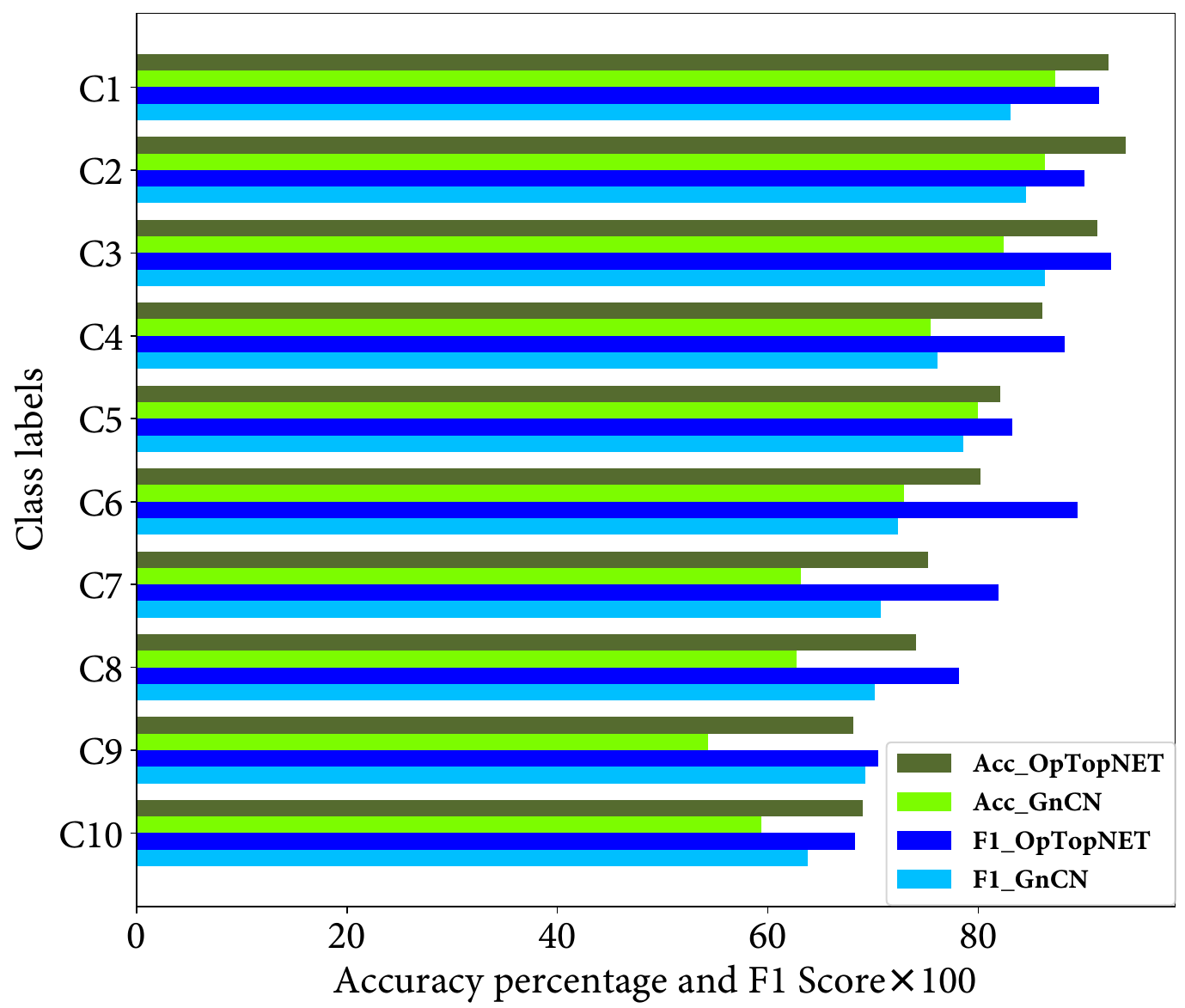}
	\caption{Average accuracy and F1 scores associated with OpTopNET and GnCN}
	\label{fig:compare}
\end{figure}

\z{We compare the prediction results of the proposed OpTopNET to the graph-normalized convolutional network (GnCN) \cite{chen2022learning}, which is the state-of-the-art method. To setup the desired GnCN, we employ adjacent-wise normalization scheme to consider the features of all robots during model training. So, the input to the graph is the adjacency matrices of various optimal topologies of a robot network that is normalized according to the rule $X_{i} \leftarrow X_{i}/|\sum_{Y \in N_{i}}Y|$, where $X_{i}$ is the coordinate of robot $i$, $N_{i}$ is the set of adjacent robots, such as $Y$, to $X_{i}$ with respect to the connection threshold define earlier, and $|\cdot|$ returns the Euclidean metric of its argument vector. The structure of the GnCN includes an input layer, an output layer, and 10 units of successive pooling-normalization layers, where average pooling in the neighborhood of each robot is taken into account. To make results comparable to the tests for OpTopNET, GnCN dimensions are set for the network of 10 robots we had already introduced. Given the same train-validation-test partitioning  approach discussed earlier for OpTopNET, we train the GnCN with batch size 16 and randomized cross-validation factor $k=10$.	To predict the optimal topology, we conduct edge-level inference to see whether or not OpTopNET results are better than those of the GnCN in prediction of the links of each optimal topology. Accordingly, Fig. \ref{fig:compare} shows that both average accuracy and F1 scores of OpTopNET for all class labels are higher than those of the GnCN.}
\vspace*{-3mm}
\section{Conclusion}
\label{sec:conc}
Optimal topology prediction for ad-hoc robot networks requires the learning and inference of complex spatial correlations between robots of a network and intended optimality criteria. In this paper, this complex problem is transformed into a multi-task classification problem in which each class is a cluster ID associated with the topology of a network. Inspired by the divide-and-conquer philosophy in algorithm design, we partition the total multi-task prediction problem corresponding to a whole robot network into a set of multi-class classification problems each of which associates with the prediction of the topological characteristics of a particular robot in the network. A stacked ensemble architecture is proposed to solve each of those partitioned multi-class classification problems. Our conducted simulations portray the effectiveness of the proposed learning methodology.

\bibliographystyle{IEEEtran} 
\bibliography{references}
\end{document}